\pdfoutput=1
\documentclass[11pt]{article}

\usepackage{emnlp2021}
\usepackage{times}
\usepackage{latexsym}
\usepackage[T1]{fontenc}
\usepackage[utf8]{inputenc}
\usepackage{microtype}
\usepackage{times}
\usepackage{latexsym}
\usepackage{multicol}

\usepackage{amsmath}
\usepackage{microtype}
\usepackage{multirow}
\usepackage{threeparttable} 
\usepackage{graphicx}

\usepackage{enumitem,kantlipsum}

\usepackage{subfigure}
\usepackage{tikz}
\usetikzlibrary{positioning} 
\usepackage{pgfplots}
\usepackage{booktabs}
\usepackage{tabularx}
\usepackage{array}
\usepackage{amsmath}
\usepackage{setspace}
\usepackage{multirow}
\usepackage{makecell}
\usepackage{xcolor}
\usepackage{colortbl}
\usepackage{amssymb}
\usepackage{MnSymbol}

\title{A Partition Filter Network for Joint Entity and Relation Extraction}

\author{Zhiheng Yan$^1$, Chong Zhang$^1$, Jinlan Fu$^{1,2}$, Qi Zhang$^{1}$\thanks{$^*$  Corresponding author.} \ and Zhongyu Wei$^3$ \\
         $^1$School of Computer Science, Shanghai Key Laboratory of Intelligent Information Processing, \\
         Fudan University, Shanghai, China \\ %
         $^2$National University of Singapore, Singapore \\
         $^3$School of Data Science, Fudan University, Shanghai, China \\
         \texttt{\{zhyan20, chongzhang20, qz, zywei\}@fudan.edu.cn} \\ 
         \texttt{jinlanjonna@gmail.com}
         }

\pgfplotsset{compat=1.17}
\begin{document}

\maketitle

\begin{abstract}
In joint entity and relation extraction, existing work either sequentially encode task-specific features, leading to an imbalance in inter-task feature interaction where features extracted later have no direct contact with those that come first. Or they encode entity features and relation features in a parallel manner, meaning that feature representation learning for each task is largely independent of each other except for input sharing. We propose a partition filter network to model two-way interaction between tasks properly, where feature encoding is decomposed into two steps: partition and filter. In our encoder, we leverage two gates: entity and relation gate, to segment neurons into two task partitions and one shared partition. The shared partition represents inter-task information valuable to both tasks and is evenly shared across two tasks to ensure proper two-way interaction. The task partitions represent intra-task information and are formed through concerted efforts of both gates, making sure that encoding of task-specific features is dependent upon each other. Experiment results on six public datasets show that our model performs significantly better than previous approaches. In addition, contrary to what previous work has claimed, our auxiliary experiments suggest that relation prediction is contributory to named entity prediction in a non-negligible way.  The source code can be found at \url{https://github.com/Coopercoppers/PFN}. 

\end{abstract}
 
\section{Introduction}
Joint entity and relation extraction intend to simultaneously extract entity and relation facts in the given text to form relational triples as ({\rm s, r, o}). The extracted information provides a supplement to many studies, such as knowledge graph construction \citep{riedel-etal-2013-relation}, question answering \citep{diefenbach2018core} and text summarization \citep{Gupta2010ASO}. 

\begin{figure}[htbp]
  \vspace{2mm}
  \centering
  \includegraphics[width=0.45\textwidth]{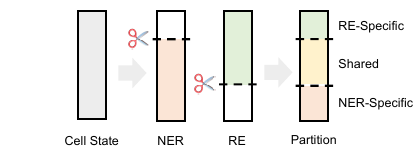}
  \caption{\label{fig: part} Partition process of cell neurons. Entity and relation gate are used to divide neurons into task-related and task-unrelated ones. Neurons relating to both tasks form the shared partition while the rest form two task partitions. }
  \vspace{-1.5mm}
\end{figure}

Conventionally, Named Entity Recognition (NER) and Relation Extraction (RE) are performed in a pipelined manner \citep{zelenko2003, chan-roth-2011-exploiting}. These approaches are flawed in that they do not consider the intimate connection between NER and RE. Also, error propagation is another drawback of pipeline methods. In order to conquer these issues, joint extracting entity and relation is proposed and demonstrates stronger performance on both tasks. In early work, joint methods mainly rely on elaborate feature engineering to establish interaction between NER and RE \citep{yu-lam-2010-jointly, li-ji-2014-incremental, miwa-sasaki-2014-modeling}. Recently, end-to-end neural network has shown to be successful in extracting relational triples \citep{zeng-etal-2014-relation, gupta-etal-2016-table, katiyar-cardie-2017-going, shen2021trigger} and has since become the mainstream of joint entity and relation extraction.

According to their differences in encoding task-specific features, most of the existing methods can be divided into two categories: sequential encoding and parallel encoding. In sequential encoding, task-specific features are generated sequentially, which means features extracted first are not affected by those that are extracted later. \citet{zeng-etal-2018-extracting} and \citet{wei-etal-2020-novel} are typical examples of this category. Their methods extract features for different tasks in a predefined order. In parallel encoding, task-specific features are generated independently using shared input. Compared with sequential encoding, models build on this scheme do not need to worry about the implication of encoding order. For example, \citet{fu-etal-2019-graphrel} encodes entity and relation information separately using common features derived from their GCN encoder. Since both task-specific features are extracted through isolated sub-modules, this approach falls into the category of parallel encoding.   

However, both encoding designs above fail to model two-way interaction between NER and RE tasks properly. In sequential encoding, interaction is only unidirectional with a specified order, resulting in different amount of information exposed to NER and RE task. In parallel encoding, although encoding order is no longer a concern, interaction is only present in input sharing. Considering adding two-way interaction in feature encoding, we adopt an alternative encoding design: joint encoding. This design encodes task-specific features jointly with a single encoder where there should exist some mutual section for inter-task communication. 

In this work, we instantiate joint encoding with a partition filter encoder. Our encoder first sorts and partitions each neuron according to its contribution to individual tasks with entity and relation gates. During this process, two task partitions and one shared partition are formed (see figure \ref{eqn:candidate cell}). Then individual task partitions and shared partition are combined to generate task-specific features, filtering out irrelevant information stored in the opposite task partition.

Task interaction in our encoder is achieved in two ways: First, the partitions, especially the task-specific ones, are formed through concerted efforts of entity and relation gates, allowing for interaction between the formation of entity and relation features determined by these partitions. Second, the shared partition, which represents information useful to both task, is equally accessible to the formation of both task-specific features, ensuring balanced two-way interaction. The contributions of our work are summarized below:

\begin{enumerate}
\vspace{-2.5mm}
\item  We propose partition filter network, a framework designed specifically for joint encoding. This method is capable of encoding task-specific features and guarantees  proper two-way interaction between NER and RE.
\vspace{-2.3mm}
\item We conduct extensive experiments on six datasets. The main results show that our method is superior to other baseline approaches, and the ablation study provides insight into what works best for our framework.
\vspace{-5mm}
\item Contrary to what previous work has claimed, our auxiliary experiments suggest that relation prediction is contributory to named entity prediction in a non-negligible way.

\end{enumerate}
\section{Related Work}
In recent years, joint entity and relation extraction approaches have been focusing on tackling triple overlapping problem and modelling task interaction. Solutions to these issues have been explored in recent works \citep{zheng-etal-2017-joint, zeng-etal-2018-extracting, zeng-etal-2019-learning, fu-etal-2019-graphrel, wei-etal-2020-novel}. The triple overlapping problem refers to triples sharing the same entity (SEO, i.e. SingleEntityOverlap) or entities (EPO, i.e. EntityPairOverlap). For example, In "Adam and Joe were born in the USA", since triples (Adam, birthplace, USA) and (Joe, birthplace, USA) share only one entity "USA", they should be categorized as SEO triples; or in "Adam was born in the USA and lived there ever since", triples (Adam, birthplace, USA) and (Adam, residence, USA) share both entities at the same time, thus should be categorized as EPO triples. Generally, there are two ways in tackling the problem. One is through generative methods like seq2seq \citep{zeng-etal-2018-extracting, zeng-etal-2019-learning} where entity and relation mentions can be decoded multiple times in output sequence, another is by modeling each relation separately with sequences \citep{wei-etal-2020-novel}, graphs \citep{fu-etal-2019-graphrel} or tables \citep{wang-lu-2020-two}. Our method uses relation-specific tables \citep{miwa-sasaki-2014-modeling} to handle each relation separately.

Task interaction modeling, however, has not been well handled by most of the previous work. In some of the previous approaches, Task interaction is achieved with entity and relation prediction sharing the same features \citep{tran2019neural, wang-etal-2020-tplinker}. This could be problematic as information about entity and relation could sometimes be contradictory. Also, as models that use sequential encoding \citep{bekoulis2018joint, eberts2019span, wei-etal-2020-novel} or parallel encoding \citep{fu-etal-2019-graphrel} lack proper two-way interaction in feature extraction, predictions made on these features suffer the problem of improper interaction. In our work, the partition filter encoder is built on joint encoding and is capable of handling communication of inter-task information more appropriately to avoid the problem of sequential and parallel encoding (exposure bias and insufficient interaction), while keeping intra-task information away from the opposite task to mitigate the problem of negative transfer between the tasks.     

\section{Problem Formulation}
Our framework split up joint entity and relation extraction into two sub-tasks: NER and RE. Formally, Given an input sequence $s = \{w_1, \dots, w_L\}$ with $L$ tokens, $w_i$ denotes the i-th token in sequence $s$. For NER, we aim to extract all typed entities whose set is denoted as $S$, where $\langle w_i, e, w_j \rangle\in S$ signifies that token $w_i$ and $w_j$ are the start and end token of an entity typed $e \in \mathcal{E}$. $\mathcal{E}$ represents the set of entity types. Concerning RE, the goal is to identify all head-only triples whose set is denoted as $T$, each triple $\langle w_i, r, w_j \rangle\in T$ indicates that tokens $w_i$ and $w_j$ are the corresponding start token of subject and object entity with relation $r \in \mathcal{R}$. $\mathcal{R}$ represents the set of relation types. Combining the results from both NER and RE, we should be able to extract relational triples with complete entity spans.  

\begin{figure*}[htbp]
  \centering
  \vspace{-3.5mm}
  \subfigure[height=0.52\textwidth][Framework of Partition Filter Network]{
  \includegraphics[height=0.52\textwidth]{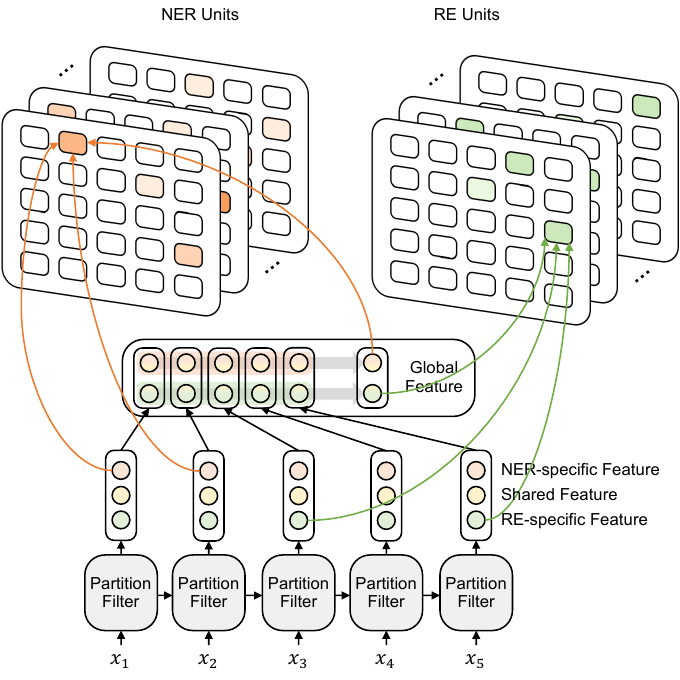}
  }
  \hspace{-0.02\textwidth}
  \subfigure[height=0.52\textwidth][Inner Mechanism of Partition Filter]{
  \includegraphics[height=0.52\textwidth]{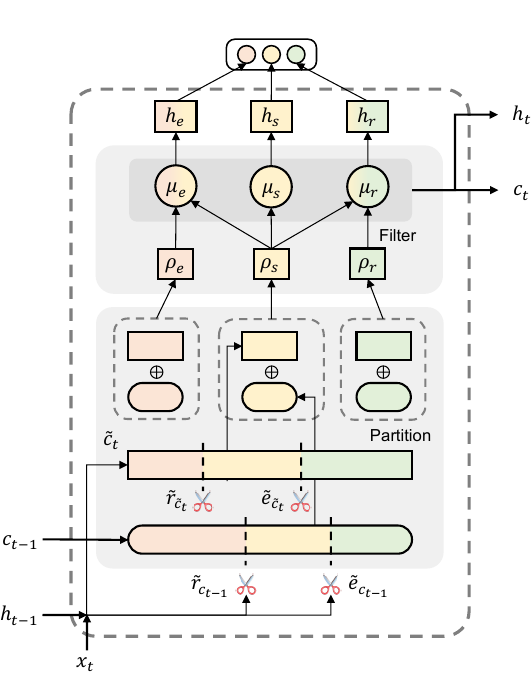}
  }
  \caption{\label{fig: model} (a) Overview of PFN. The framework consists of three components: partition filter encoder, NER unit and RE unit. In task units, we use table-filling for word pair prediction. Orange, yellow and green represents NER-related, shared and RE-related component or features. (b) Detailed depiction of partition filter encoder in one single time step. We decompose feature encoding into two steps: partition and filter (shown in the gray area). In partition, we first segment neurons into two task partitions and one shared partition. Then in filter, partitions are selected and combined to form task-specific features and shared features, filtering out information irrelevant to each task. }
  \vspace{-1.5mm}
\end{figure*}

\section{Model}
We describe our model design in this section. Our model consists of a partition filter encoder and two task units, namely NER unit and RE unit. The partition filter encoder is used to generate task-specific features, which will be sent to task units as input for entity and relation prediction. We will discuss each component in detail in the following three sub-sections.

\subsection{Partition Filter Encoder} 
Similar to LSTM, the partition filter encoder is a recurrent feature encoder with information stored in intermediate memories. In each time step, the encoder first divides neurons into three partitions: entity partition, relation partition and shared partition. Then it generates task-specific features by selecting and combining these partitions, filtering out information irrelevant to each task. As shown in figure \ref{fig: model}, this module is designed specifically to jointly extract task-specific features, which strictly follows two steps: partition and filter. 

\paragraph{Partition} 
This step performs neuron partition to divide cell neurons into three partitions: Two task partitions storing intra-task information, namely entity partition and relation partition, as well as one shared partition storing inter-task information. The neuron to be divided are candidate cell $\tilde{c}_t$ representing current information and previous cell $c_{t-1}$ representing history information. $c_{t-1}$ is the direct input from the last time step and $\tilde{c}_t$ is calculated in the same manner as LSTM:

\begin{equation}\label{eqn:candidate cell} 
\tilde{c}_t = \tanh({\rm{Linear}} ([x_t; h_{t-1}]))
\end{equation}
where $\rm{Linear}$ stands for the operation of linear transformation. 

We leverage entity gate $\tilde{e}$ and relation gate $\tilde{r}$, which are referred to as master gates in  \citep{shen2018ordered}, for neuron partition. As illustrated in figure \ref{eqn:candidate cell}, each gate, which represents one specific task, will divide neurons into two segments according to their usefulness to the designated task. For example, entity gate $\tilde{e}$ will separate neurons into two partitions: NER-related and NER-unrelated. The shared partition is formed by combining partition results from both gates. Neurons in the shared partition can be regarded as information valuable to both tasks. In order to model two-way interaction properly, inter-task information in the shared partition is evenly accessible to both tasks (which will be discussed in the filter subsection). In addition, information valuable to only one task is invisible to the opposing task and will be stored in individual task partitions.
The gates are calculated using cummax activation function $cummax\left(\cdot\right)=cumsum(softmax(\cdot))$\footnotemark, whose output can be seen as approximation of a binary gate with the form of $(0,\dots ,0,1,\dots ,1)$: 
\footnotetext{$cumsum(x_1,x_2,\dots,x_{n-1},x_n)=(x_1,x_1+x_2,\dots,\\
x_1+x_2+\dots+x_{n-1},x_1+x_2+\dots +x_{n-1}+x_n)$.}
\begin{equation}\label{eqn:3} 
\begin{array}{l@{}l} 
{\tilde{e} ={\rm cummax}({\rm{Linear}} ([x_{t};h_{t-1}]))} \\ {\tilde{r} =1-{\rm cummax}({\rm{Linear}} ([x_{t};h_{t-1}]))} \end{array}
\end{equation}
The intuition behind equation (\ref{eqn:3}) is to identify two cut-off points, displayed as scissors in figure \ref{fig: model}, which naturally divide a set of neurons into three segments.

As a result, the gates will divide neurons into three partitions, entity partition $\rho_e$, relation partition $\rho_r$ and shared partition $\rho_s$. Partitions for previous cell $c_{t-1}$ are formulated as below:  \footnote{The calculation for candidate cell $\tilde{c}_t$ is practically identical to equation (\ref{eqn:5}) and therefore not shown.}
\begin{equation}\label{eqn:5}
\begin{array}{l@{}l} 
{\rho_{s, c_{t-1}}  =\tilde{e}_{c_{t-1}} \circ \tilde{r}_{c_{t-1}} } \\
\rho_{e, c_{t-1}} = \tilde{e}_{c_{t-1}} - \rho_{s, c_{t-1}} \\ 
\rho_{r, c_{t-1}} = \tilde{r}_{c_{t-1}} - \rho_{s, c_{t-1}}  \end{array} 
\end{equation}
Note that if you add up all three partitions, the result is not equal to one. This guarantees that in forward message passing, some information is discarded to ensure that message is not overloaded, which is similar to the forgetting mechanism in LSTM. 

Then, we aggregate partition information from both target cells, and three partitions are formed as a result. For all three partitions, we add up all related information from both cells:
\begin{equation}
\begin{array}{l@{}l}
{\rho_{e}=\rho_{e, c_{t-1}} \circ c_{t-1} + \rho_{e, \tilde{c}_{t}} \circ \tilde{c}_{t}} \\
{\rho_{r}=\rho_{r, c_{t-1}} \circ c_{t-1} + \rho_{r, \tilde{c}_{t}} \circ \tilde{c}_{t}} \\
{\rho_{s} =  \rho_{s, c_{t-1}} \circ c_{t-1}  + \rho_{s, \tilde{c}_{t}} \circ \tilde{c}_{t} }
\end{array}
\end{equation}
\paragraph{Filter} We propose three types of memory block: entity memory, relation memory and shared memory. Here we denote $\mu_{e}$ as entity memory, $\mu_{r}$ as relation memory and $\mu_{s}$ as shared memory. In $\mu_{e}$, information in entity partition and shared partition are selected. In contrast, information in relation partition, which we assume is irrelevant or even harmful to named entity recognition task, is filtered out. The same logic applies to $\mu_{r}$ as well, where information in entity partition is filtered out and the rest is kept. In addition, information in shared partition will be stored in $\mu_{s}$:
\begin{equation}
\begin{array}{l@{}l} 
{\mu_{e} =\rho_{e} + \rho_{s}}; \
{\mu_{r} =\rho_{r} + \rho_{s}}; \
{\mu_{s} =\rho_{s} }
\end{array}
\end{equation}
Note that inter-task information in the shared partition is accessible to both entity memory and relation memory, allowing balanced interaction between NER and RE. Whereas in sequential and parallel encoding, relation features have no direct impact on the formation of entity features.

After updating information in each memory, entity features $h_{e}$, relation features $h_{r}$ and shared features $h_{s}$ are generated with corresponding memories:
\begin{equation}
\begin{array}{l@{}l} 
{h_{e} =\tanh (\mu_{e})} \\ 
{h_{r} =\tanh (\mu_{r})} \\
{h_{s} =\tanh (\mu_{s})} 
\end{array}
\end{equation}
Following the partition and filter steps, information in all three memories is used to form cell state $c_t$,  which will then be used to generate hidden state $h_{t}$ (The hidden and cell state at time step $t$ are input to the next time step):
\begin{equation}
\begin{array}{l@{}l} 
{c_{t} = {\rm{Linear}} ([\mu_{e,t}; \mu_{r,t}; \mu_{s,t}])  } \\ 
{h_{t} = \tanh (c_t) } 
\end{array}
\end{equation}

\subsection{Global Representation} In our model, we employ a unidirectional encoder for feature encoding. The backward encoder in the bidirectional setting is replaced with task-specific global representation to capture the semantics of future context. Empirically this shows to be more effective. For each task, global representation is the combination of task-specific features and shared features computed by:    

\begin{equation} 
\begin{array}{l@{}l} 
{h_{g_{e},t} = {\tanh(\rm{Linear}} [h_{e,t}; h_{s,t}])} \\
{h_{g_{r},t} = {\tanh(\rm{Linear}} [h_{r,t}; h_{s,t}])} \\
{h_{g_{e}} ={\rm maxpool}(h_{g_{e},1} ,\dots,h_{g_{e}, L}) } \\
{h_{g_{r}} ={\rm maxpool}(h_{g_{r},1} ,\dots,h_{g_{r}, L}) }
\end{array}
\end{equation}

\subsection{Task Units}
Our model consists of two task units: NER unit and RE unit. In NER unit, the objective is to identify and categorize all entity spans in a given sentence. More specifically, the task is treated as a type-specific table filling problem. Given a entity type set $\mathcal{E}$, for each type $k$, we fill out a table whose element $e^k_{ij}\ $ represents probability of word $w_i$ and word $w_j$ being start and end position of an entity with type $k$.  For each word pair$\ (w_i,w_j)$, we concatenate word-level entity features $h^{e}_{i}$ and $h^{e}_{j}$, as well as sentence-level global features $h_{g_{e}}$ before feeding it into a fully-connected layer with ELU activation to get entity span representation $h^{e}_{ij}$:
\begin{equation}
h^{e}_{ij} ={\rm ELU} ({\rm{Linear}} ([h^{e}_{i};h^{e}_{j};h_{g_{e}} ]))
\end{equation}
With the span representation, we can predict whether the span is an entity with type $k$ by feeding it into a feed forward neural layer:
\begin{equation}
\begin{array}{l@{}l}
\begin{aligned}
e_{ij}^k &=p(e=\left \langle w_{i},k,w_{j}  \right \rangle|e\in S) \\ 
&=\sigma({\rm{Linear}}(h^{e}_{ij})), \forall k \in \mathcal{E}
\end{aligned}
\end{array}
\end{equation}
where $\sigma$ represents sigmoid activation function.

Computation in RE unit is mostly symmetrical to NER unit. Given a set of gold relation triples denoted as $T$, this unit aims to identify all triples in the sentence. We only predict starting word of each entity in this unit as entity span prediction is already covered in NER unit. Similar to NER, we consider relation extraction as a relation-specific table filling problem. Given a relation label set $\mathcal{R}$, for each relation $l\in \mathcal{R}$, we fill out a table whose element $r^l_{ij}$ represents the probability of word $w_i$ and word $w_j$ being starting word of subject and object entity. In this way, we can extract all triples revolving around relation $l$ with one relation table. For each triple $(w_i,l, w_j)$, similar to NER unit, triple representation $h^{r}_{ij}$ and relation score $r^l_{ij}$ are calculated as follows:
\begin{equation}
\begin{aligned}[rl]
h^{r}_{ij} &={\rm ELU}({\rm{Linear}}([h^{r}_{i}; h^{r}_{j};h_{g_{r}} ])) \\
r_{ij}^{l} &=p(r=\left \langle w_{i},l,w_{j} \right \rangle|r\in T) \\
&=\sigma ({\rm{Linear}}(h^{r}_{ij})), \forall l \in \mathcal{R}
\end{aligned}
\end{equation}

\subsection{Training and Inference}
For a given training dataset, the loss function $L\ $that guides the model during training consists of two parts:${\ L}_{ner}$ for NER unit and $L_{re}$ for RE unit:
\begin{equation}
\begin{array}{l@{}l} 
L_{ner} = \sum\limits_{\hat{e}_{ij}^k\in S}{{\rm{BCELoss}}(e_{ij}^k, \hat{e}_{ij}^k)} \\ 
L_{re} = \sum\limits_{\hat{r}_{ij}^l\in T}{{\rm{BCELoss}}(r_{ij}^l, \hat{r}_{ij}^l)} \\
\end{array}        
\end{equation} 
$\hat{e}_{ij}^k$ and $\hat{r}_{ij}^l$ are respectively ground truth label of entity table and relation table. $e_{ij}^k$ and $r_{ij}^l$ are the predicted ones. We adopt $\rm{BCELoss}$ for each task\footnote{${\rm{BCELoss}}(x,y)=-(y{\rm{log}}x+(1-y){\rm{log}}(1-x)).$}. The training objective is to minimize the loss function $L$, which is computed as $L_{ner} +L_{re}$.

During inference, we extract relational triples by combining results from both NER and RE unit. For each legitimate triple prediction $(s^{k}_{i, j},l,o^{k'}_{m, n})$ where $l$ is the relation label, $k$ and $k'$ are the entity type labels, and the indexes $i, j$ and $m, n$ are respectively starting and ending index of subject entity $s$ and object entity $o$, the following conditions should be satisfied:

\begin{equation}
\begin{array}{l@{}l} {e^{k}_{ij} \ge \lambda _{e} };\ {e^{k'}_{mn} \ge \lambda _{e} };\  {r_{im}^{l} \ge \lambda _{r} }
\end{array}  
\end{equation}
${\lambda }_e$ and ${\lambda }_r$ are threshold hyper-parameters for entity and relation prediction, both set to be 0.5 without further fine-tuning.

\section{Experiment}

\subsection{Dataset, Evaluation and Implementation Details}
We evaluate our model on six datasets. NYT \citep{riedel2010modeling}, WebNLG \citep{zeng-etal-2018-extracting}, ADE \citep{gurulingappa2012development}, SciERC \citep{luan-etal-2018-multi}, ACE04 and ACE05 \citep{walker2006ace}. Descriptions of the datasets can be found in Appendix \ref{sec A}.

Following previous work, we assess our model on NYT/WebNLG under \textbf{partial match}, where only the tail of an entity is annotated. Besides, as entity type information is not annotated in these datasets, we set the type of all entities to a single label "NONE", so entity type would not be predicted in our model. On ACE05, ACE04, ADE and SciERC, we assess our model under \textbf{exact match} where both head and tail of an entity are annotated. For ADE and ACE04, 10-fold and 5-fold cross validation are used to evaluate the model respectively, and 15\% of the training set is used to construct the development set. For evaluation metrics, we report F1 scores in both NER and RE. In NER, an entity is seen as correct only if its type and boundary are correct. In RE, A triple is correct only if the types, boundaries of both entities and their relation type are correct. In addition, we report Macro-F1 score in ADE and Micro-F1 score in other datasets. 

We choose our model parameters based on the performance in the development set (the best average F1 score of NER and RE) and report the results on the test set. More details of hyper-parameters can be found in Appendix \ref{sec: Appendix c}

\begin{table}[!ht]
\footnotesize
\centering
\begin{spacing}{1.19}
\setlength{\tabcolsep}{0.8mm}{
\begin{tabular}{lcc}
\toprule
\textbf{Method} & NER & RE\\
\midrule
\multicolumn{3}{l}{\textbf{NYT $\vartriangle$}} \\
CopyRE \text{ \citep{zeng-etal-2018-extracting}} & 86.2 & 58.7 \\
GraphRel \text{\ \citep{fu-etal-2019-graphrel}} & 89.2 & 61.9 \\
CopyRL \citep{zeng-etal-2019-learning} & - & 72.1 \\
Casrel \text{\ \citep{wei-etal-2020-novel}} $^{\dag}$ & (93.5) & 89.6 \\ 
TpLinker \citep{wang-etal-2020-tplinker} $^{\dag}$ & - & 91.9 \\
\hline
PFN$^{\dag}$ & \textbf{95.8} & \textbf{92.4} \\
\midrule
\multicolumn{3}{l}{\textbf{WebNLG $\vartriangle$}} \\
CopyRE \text{ \citep{zeng-etal-2018-extracting}} & 82.1 & 37.1 \\
GraphRel \text{\ \citep{fu-etal-2019-graphrel}} & 91.9 & 42.9 \\
CopyRL \citep{zeng-etal-2019-learning} & - & 61.6 \\
Casrel \text{\ \citep{wei-etal-2020-novel}} $^{\dag}$ & (95.5) & 91.8 \\ 
TpLinker \citep{wang-etal-2020-tplinker} $^{\dag}$ & - & 91.9 \\
\hline
PFN$^{\dag}$ & \textbf{98.0} & \textbf{93.6} \\
\midrule
\multicolumn{3}{l}{\textbf{ADE $\blacktriangle$}} \\
Multi-head \citep{bekoulis2018joint} & 86.4 & 74.6 \\
Multi-head + AT \citep{bekoulis2018adversarial} & 86.7 & 75.5 \\
Rel-Metric \citep{tran2019neural} & 87.1 & 77.3 \\
SpERT \citep{eberts2019span} $^{\dag}$ & 89.3 & 79.2 \\
Table-Sequence \citep{wang-lu-2020-two} $^{\ddag}$ & 89.7 & 80.1 \\
\hline
PFN$^{\dag}$ & 89.6 & 80.0 \\
PFN$^{\ddag}$ & \textbf{91.3} & \textbf{83.2} \\
\midrule
\multicolumn{3}{l}{\textbf{ACE05} $\vartriangle$} \\
Structured Perceptron \citep{li-ji-2014-incremental} & 80.8 & 49.5 \\ 
SPTree \citep{miwa-bansal-2016-end} & 83.4 & 55.6 \\
Multi-turn QA \citep{li-etal-2019-entity} $^{\dag}$ & 84.8 & 60.2 \\
Table-Sequence \citep{wang-lu-2020-two} $^{\ddag}$ & 89.5 & 64.3 \\
PURE \citep{zhong2021frustratingly} $^{\ddag}$ & \textbf{89.7} & 65.6 \\
\hline
PFN$^{\ddag}$ & 89.0 & \textbf{66.8} \\
\midrule
\multicolumn{3}{l}{\textbf{ACE04} $\vartriangle$} \\
Structured Perceptron \citep{li-ji-2014-incremental} & 79.7 & 45.3 \\ 
SPTree \citep{miwa-bansal-2016-end} & 81.8 & 48.4 \\
Multi-turn QA \citep{li-etal-2019-entity} $^{\dag}$ & 83.6 & 49.4 \\
Table-Sequence \citep{wang-lu-2020-two} $^{\ddag}$ & 88.6 & 59.6 \\
PURE \citep{zhong2021frustratingly} $^{\ddag}$ & 88.8 & 60.2 \\
\hline
PFN$^{\ddag}$ & \textbf{89.3} & \textbf{62.5} \\
\midrule
\multicolumn{3}{l}{\textbf{SciERC} $\vartriangle$} \\
SPE \citep{wang-etal-2020-pre} $^{\S}$ & \textbf{68.0} & 34.6 \\
PURE \citep{zhong2021frustratingly} $^{\S}$ & 66.6 & 35.6 \\
\hline
PFN$^{\S}$ &66.8 &\textbf{38.4} \\
\bottomrule
\end{tabular}
}
\end{spacing}
\caption{Experiment results on six datasets. $^{\dag}$, $^{\ddag}$ and $^{\S}$ denotes the use of BERT, ALBERT and SCIBERT\citep{devlin2019bert, Lan2020ALBERT:, beltagy-etal-2019-scibert} pre-trained embedding. $\vartriangle$ and $\blacktriangle$ denotes the use of micro-F1 and macro-F1 score. NER results of Casrel are its reported average score of head and tail entity. Results of PURE are reported in single-sentence setting for fair comparison.}
\label{tab:my-table2}
\end{table}

\subsection{Main Result}
Table \ref{tab:my-table2} shows the comparison of our model with existing approaches. In partially annotated datasets WebNLG and NYT, under the setting of BERT. For RE, our model achieves 1.7\% improvement in WebNLG but performance in NYT is only slightly better than previous SOTA TpLinker \citep{wang-etal-2020-tplinker} by 0.5\% margin. We argue that this is because NYT is generated with distant supervision, and annotation for entity and relation are often incomplete and wrong. Compared to TpLinker, the strength of our method is to reinforce two-way interaction between entity and relation. However, when dealing with noisy data, the strength might be counter-productive as error propagation between both tasks is amplified as well. 

For NER, our method shows a distinct advantage over baselines that report the figures. Compared to Casrel \citep{wei-etal-2020-novel}, a competitive method, our F1 scores are 2.3\%/2.5\% higher in NYT/WebNLG. This proves that exposing relation information to NER, which is not present in Casrel, leads to better performance in entity recognition.

Furthermore, our model demonstrates strong performance in fully annotated datasets ADE, ACE05, ACE04 and SciERC. For ADE, our model surpasses table-sequence \citep{wang-lu-2020-two} by 1.6\%/3.1\% in NER/RE. For ACE05, our model surpasses PURE \citep{zhong2021frustratingly} by 1.2\% in RE but results in weaker performance in NER by 0.7\%. We argue that it could be attributed to the fact that, unlike the former three datasets, ACE05 contains many entities that do not belong to any triple. Thus utilizing relation information for entity prediction might not be as fruitful as that in other datasets (PURE is a pipeline approach where relation information is unseen to entity prediction). In ACE04, our model surpasses PURE by 0.5\%/2.3\% in NER/RE. In SciERC, our model surpasses PURE by 0.2\%/2.8\% in NER/RE. Overall, the performance of our model shows remarkable improvement against previous baselines.

\subsection{Ablation Study}
In this section, we take a closer look and check the effectiveness of our framework in relation extraction concerning five different aspects: number of encoder layer, bidirectional versus unidirectional, encoding scheme, partition granularity and decoding strategy.  

\paragraph{Number of Encoder Layers}
Similar to recurrent neural network, we stack our partition filter encoder with an arbitrary number of layers. Here we only examine frameworks with no more than three layers. As shown in table \ref{tab:my-table3}, adding layers to our partition filter encoder leads to no improvement in F1-score. This shows that one layer is good enough for encoding task-specific features. 

\paragraph{Bidirection Vs Unidirection}
Normally we need two partition filter encoders (one in reverse order) to model interaction between forward and backward context. However, as discussed in section 4.2, our model replaces the backward encoder with a global representation to let future context be visible to each word, achieving a similar effect with bidirectional settings. In order to find out which works best, we compare these two methods in our ablation study. From table \ref{tab:my-table3}, we find that unidirectional encoder with global representation outperforms bidirectional encoder without global representation, showing that global representation is more suitable in providing future context for each word than backward encoder. In addition, when global representation is involved, unidirectional encoder achieves similar result in F1 score compared to bidirectional encoder, indicating that global representation alone is enough in capturing semantics of future context.

\paragraph{Encoding Scheme}
We replace our partition filter encoder with two LSTM variants to examine the effectiveness of our encoder. In the parallel setting, we use two LSTM encoders
to learn task-specific features separately, and no interaction is allowed except for sharing the same input. In the sequential setting where only one-way interaction is allowed, entity features generated from the first LSTM encoder is fed into the second one to produce relation features. From table \ref{tab:my-table3}, we observe that our partition filter outperforms LSTM variants by a large margin, proving the effectiveness of our encoder in modelling two-way interaction over the other two encoding schemes.

\definecolor{mygray}{gray}{.9}
\begin{table}[!t]
\small
\centering
\begin{spacing}{1.19}
\begin{threeparttable}
\begin{tabular}{c|lccc}
\bottomrule
\textbf{Ablation} & \textbf{Settings} & P & R & F \\
\hline
\multirow{3}{*}{Layers} & 
N=1 & \textbf{40.6} & \textbf{36.5} & \textbf{38.4} \\
& N=2 & 39.9 & 35.7 & 37.7 \\
& N=3 & 40.0 & 36.2 & 38.0 \\ 
\hline
\multirow{4}{*}{\makecell[c]{Bidirection \\ Vs \\ Unidirection}} & Unidirection & \textbf{40.6} & \textbf{36.5} & \textbf{38.4} \\
& \ \ (w/o gl.) & 40.5 & 34.6 & 37.3 \\
& Bidirection & 40.4 & 36.2 & 38.2\\ 
& \ \ (w/o gl.) & 39.9 & 35.3 & 37.5 \\ 
\hline
\multirow{3}{*}{\makecell[c]{Encoding \\ Scheme}} & Joint & \textbf{40.6} & \textbf{36.5} & \textbf{38.4} \\
& Sequential & 40.0 & 34.2 & 36.9 \\
& Parallel & 36.0 & 34.4 & 35.1 \\ \hline
\multirow{2}{*}{\makecell[c]{Partition \\ Granularity}} & Fine-grained & \textbf{40.6} & \textbf{36.5} & \textbf{38.4} \\
& Coarse & 39.3 & 35.5 & 37.3 \\ 
\hline
\multirow{2}{*}{\makecell[c]{Decoding \\ Strategy}} &Universal & \textbf{40.6} & \textbf{36.5} & \textbf{38.4} \\
& Selective & 38.5 & 36.3 & 37.4 \\
\toprule
\end{tabular} 
\end{threeparttable}
\end{spacing}
\caption{Ablation study on SciERC. P, R and F represent precision, recall and F1 relation scores. The best results are marked in bold. gl. in the second experiment is short for global representation. }
\label{tab:my-table3}
\end{table}

\paragraph{Partition Granularity}
Similar to \citep{shen2018ordered}, we split neurons into several chunks and perform partition within each chunk. Each chunk shares the same entity gate and relation gate. Thus partition results for all chunks remain the same. For example, with a 300-dimension neuron set, if we split it into 10 chunks, each with 30 neurons, only two 30-dimension gates are needed for neuron partition. We refer to the above operation as coarse partition. In contrast, our fine-grained partition can be seen as a special case as neurons are split into only one chunk.  We compare our fine-grained partition (chunk size = 300) with coarse partition (chunk size = 10). Table \ref{tab:my-table3} shows that fine-grained partition performs better than coarse partition. It is not surprising as in coarse partition, the assumption of performing the same neuron partition for each chunk might be too strong for the encoder to separate information for each task properly. 

\paragraph{Decoding Strategy}
In pipeline-like methods, relation prediction is performed on entities that the system considers as valid in their entity prediction. We argue that a better way for relation prediction is to take into account all the invalid word pairs. We refer to the former strategy as selective decoding and the latter one as universal decoding. For selective decoding, we only predict the relation scores for entities deemed as valid by their entity scores calculated in the NER unit.  Table \ref{tab:my-table3} shows that universal decoding, where all the negative instances are included, is better than selective decoding. Apart from mitigating error propagation, we argue that universal decoding is similar to contrastive learning as negative instances helps to better identify the positive instances through implicit comparison.

\begin{table}[!t]
\small
\centering
\begin{spacing}{1.19}
\begin{threeparttable}
\begin{tabular}{c|lcccc}
\bottomrule
\textbf{Dataset} & \textbf{Entity Type} & P & R & F & Ratio\\
\hline
\multirow{5}{*}{ACE05} 
& Total & 89.3 & 88.8 & 89.0 & 1.00\\ 
& In-triple & 95.9 & 92.1 & 94.0 & 0.36\\
& Out-of-triple & 85.8 & 86.9 & 86.3 & 0.64\\
& Diff & 10.1 & 5.2 & 7.7 & -\\
\hline
\multirow{5}{*}{ACE04} 
& Total & 89.1 & 89.6 & 89.3 & 1.00 \\ 
& In-triple & 94.3 & 91.2 & 92.7 & 0.71 \\
& Out-of-triple & 87.1 & 89.2 & 88.1 & 0.29\\
& Diff & 7.2 & 3.0 & 4.6 & - \\
\hline
\multirow{5}{*}{SciERC} 
& Total & 64.8 & 69.0 & 66.8 & 1.00\\ 
& In-triple & 78.0 & 71.1 & 74.4 & 0.78\\
& Out-of-triple & 38.9 & 61.7 & 47.8 & 0.22\\
& Diff & 39.1 & 9.4 & 26.6 & -\\
\toprule
\end{tabular} 
\end{threeparttable}
\end{spacing}
\caption{NER Results on different entity types. Entities are split into two groups: In-triple and Out-of-triple based on whether they appear in relational triples or not. Diff is the performance difference between In-triple and Out-of-triple. Ratio is number of entities of given type divided by number of total entities in the test set (train, dev and test set combined in ACE04). Results of ACE04 are averaged over 5-folds}
\label{tab:my-table4}
\end{table}

\section{Effects of Relation Signal on Entity Recognition}
It is a widely accepted fact that entity recognition helps in predicting relations, but the effect of relation signals on entity prediction remains divergent among researchers.

Through two auxiliary experiments, we find that the absence of relation signals has a considerable bearing on entity recognition.

\begin{table*}[t]
\scriptsize
\centering
\begin{spacing}{1.2}
\setlength{\tabcolsep}{1mm}{
\begin{tabular}{l|cc|cc|cc|cc|cc|c}
\bottomrule
\multirow{2}{*}{\textbf{Model}} & \multicolumn{2}{c|}{\textbf{ConcatSent}} & \multicolumn{2}{c|}{\textbf{CrossCategory}} & \multicolumn{2}{c|}{\textbf{EntTypos}} & \multicolumn{2}{c|}{\textbf{OOV}} & \multicolumn{2}{c|}{\textbf{SwapLonger}} & \textbf{Average} \\
 & Ori $\to$ Aug & Decline & Ori $\to$ Aug & Decline & Ori $\to$ Aug & Decline & Ori $\to$ Aug & Decline & Ori $\to$ Aug & Decline & Decline \\
\hline
BiLSTM-CRF & 83.0$\to$82.2 & 0.8 & 82.9$\to$43.5 & 39.4 & 82.5$\to$73.5 & 9.0 & 82.9$\to$64.2 & 18.7 & 82.9$\to$67.7 & 15.2 & 16.6 \\
BERT-base(cased) & 87.3$\to$86.2 & 1.1 & 87.4$\to$48.1 & 39.3 & 87.5$\to$83.1 & 4.1 & 87.4$\to$79.0 & 8.4 & 87.4$\to$82.1 & 5.3 & 11.6 \\
BERT-base(uncased) & 88.8$\to$88.7 & \textbf{0.1} & 88.7$\to$46.0 & 42.7 & 89.1$\to$83.0 & 6.1 & 88.7$\to$74.6 & 14.1 & 88.7$\to$78.5 & 10.2 & 14.6 \\
TENER & 84.2$\to$83.4 & 0.8 & 84.7$\to$39.6 & 45.1 & 84.5$\to$76.6 & 7.9 & 84.7$\to$51.5 & 33.2 & 84.7$\to$31.1 & 53.6 & 28.1 \\
Flair & 85.5$\to$85.2 & 0.3 & 84.6$\to$44.9 & 39.7 & 86.1$\to$81.5 & 4.6 & 84.6$\to$81.3 & \textbf{3.3} & 84.6$\to$73.1 & 11.5 & 11.9 \\
\hline
PFN & 89.1$\to$87.9 & 1.2 & 89.0$\to$80.5 & \textbf{8.5} & 89.6$\to$86.9 & \textbf{2.7} & 89.0$\to$80.4 & 8.6 & 89.0$\to$84.3 & \textbf{4.7} & \textbf{5.1} \\
\toprule
\end{tabular}}
\end{spacing}
\caption{Robustness test of NER against input perturbation in ACE05, baseline results and test files are copied from \url{https://www.textflint.io/}}
\label{tab:textflint}
 \end{table*}
 
\subsection{Analysis on Entity Prediction of Different Types}
\label{6-1}
In Table \ref{tab:my-table2}, NER performance of our model is consistently better than other baselines except for ACE05 where the performance falls short with a non-negligible margin. We argued that it could be attributed to the fact that ACE05 contains many entities that do not belong to any triples. 

To corroborate our claim, in this section we try to quantify the performance gap of entity prediction between entities that belong to certain triples and those that have no relation with other entities. The former ones are referred to as In-triple entities and the latter as Out-of-triple entities. We split the entities into two groups and test the NER performance of each group in ACE05/ACE04/SciERC. In NYT/WebNLG/ADE, since Out-of-triple entity is non-existent, evaluation is not performed on these datasets.   

As is shown in table \ref{tab:my-table4}, there is a huge gap between In-triple entity prediction and Out-of-triple entity prediction, especially in SciERC where the diff score reaches 26.6\%. We argue that it might be attributed to the fact that entity prediction in SciERC is generally harder given that it involves identification of scientific terms and also the average length of entities in SciERC are longer. Another observation is that the diff score is largely attributed to the difference of precision, which means that without guidance from relational signal, our model tends to be over-optimistic about entity prediction.

In addition, compared to PURE \citep{zhong2021frustratingly} we find that the overall performance of NER is negatively correlated with the percentage of out-of-triple entities in the dataset. especially in ACE05, where the performance of our model is relatively weak, over 64\% of the entities are Out-of-triple. This phenomenon is a manifest of the weakness in joint model: Joint modeling of NER and RE might be somewhat harmful to entity prediction as the inference patterns of In-triple and Out-of-triple entity are different, considering that the dynamic between relation information and entity prediction is different for In-triple and Out-of-triple entity.

\subsection{Robustness Test on Named Entity Recognition}
\label{6-2}
We use robustness test to evaluate our model under adverse circumstances. In this case, we use the domain transformation methods of NER from \citep{wang2021textflint}. The compared baselines are all relation-free models, including BiLSTM-CRF \citep{huang2015bidirectional}, BERT \citep{devlin2019bert}, TENER \citep{yan2019tener} and Flair-Embeddings \citep{akbik2019flair}. Descriptions of the transformation methods can be found in Appendix \ref{sec B}

From table \ref{tab:textflint}, we observe that our model is mostly more resilient against input perturbations compared to other baselines, especially in the category of CrossCategory, which is probably attributed to the fact that relation signals used in our training impose type constraints on entities, thus inference of entity types is less affected by the semantic meaning of target entity itself, but rather the (relational) context surrounding the entity.

\subsection{Does Relation Signal Helps in Predicting Entities}
Contrary to what \citep{zhong2021frustratingly} has claimed (that relation signal has minimal effects on entity prediction), we find several clues that suggest otherwise. First, in section \ref{6-1}, we observe that In-triple entities are much more easier to predict than Out-of-triple entities, which suggests that relation signals are useful to entity prediction. Second,
in section \ref{6-2}, we perform robustness test in NER to evaluate our model's capability against input perturbation. In the robustness test we compare our method - the only joint model to other relation-free baselines. The result suggests that our method is much more resilient against adverse circumstances, which could be (at least partially) explained by the introduction of relation signals.
To sum up, we find that relation signals do have non-negligible effect on entity prediction. The reason for \citep{zhong2021frustratingly} to conclude that relation information has minimal influence on entity prediction is most probably due to selective bias, meaning that the evaluated dataset ACE05 contains a large proportion of Out-of-triple entities (64\%), which in essence does not require any relation signal themselves.

\section{Conclusion}
In this paper, we encode task-specific features with our newly proposed model: Partition Filter Network in joint entity and relation extraction. Instead of extracting task-specific features in a sequential or parallel manner, we employ a partition filter encoder to generate task-specific features jointly in order to model two-way inter-task interaction properly. We conduct extensive experiments on six datasets to verify the effectiveness of our model. Overall experiment results demonstrate that our model is superior to previous baselines in entity and relation prediction. Furthermore, dissection on several aspects of our model in ablation study sheds some light on what works best in our framework.  Lastly, contrary to what previous work has claimed, our auxiliary experiments suggest that relation prediction is contributory to named entity prediction in a non-negligible way.

\section{Acknowledgements}
The authors wish to thank the anonymous reviewers for their helpful comments. This work was partially funded by China National Key R\&D Program (No.2018YFB1005104), National Natural Science Foundation of China (No.62076069, 61976056), Shanghai Municipal Science and Technology Major Project (No.2021SHZDZX0103).

\bibliographystyle{acl_natbib}
\bibliography{anthology,custom}


\appendix

\newcolumntype{Y}{C{0.05\textwidth}}
\begin{table}[]
\small
\centering
\begin{spacing}{1.19}
\begin{threeparttable}
\begin{tabular}{lccccc}
\bottomrule
\multirow{2}{*}{\textbf{Dataset}} & \multicolumn{3}{c}{\textbf{\#Sentences}} & \multirow{2}{*}{$|\mathcal{E}|$} & \multirow{2}{*}{$|\mathcal{R}|$} \\
 & Train & Dev & Test & & \\
\hline
NYT & 56,195 & 5,000 & 5,000 & - & 24 \\
WebNLG & 5,019 & 500 & 703 & - & 170 \\
ADE & \multicolumn{3}{c}{4,272\ (10-fold)} & 2 & 1 \\
ACE05 & 10,051 & 2,424 & 2,050 & 7 & 6 \\
ACE04 & \multicolumn{3}{c}{8,683\ (5-fold)} & 7 & 6 \\
SciERC & 1,861 & 275 & 551 &6 &7 \\
\toprule
\end{tabular} 
\end{threeparttable}
\end{spacing}
\caption{Statistics of datasets. $|\mathcal{E}|$ and $|\mathcal{R}|$ are numbers of entity and relation types. In NYT and WebNLG, entity type information is not annotated.}
\label{tab:my-table1}
\end{table}

\begin{figure*}[htbp]
  \vspace{-0.2cm}
  \centering
  \subfigure[width=0.49\textwidth][NYT]{
  \includegraphics[width=0.49\textwidth]{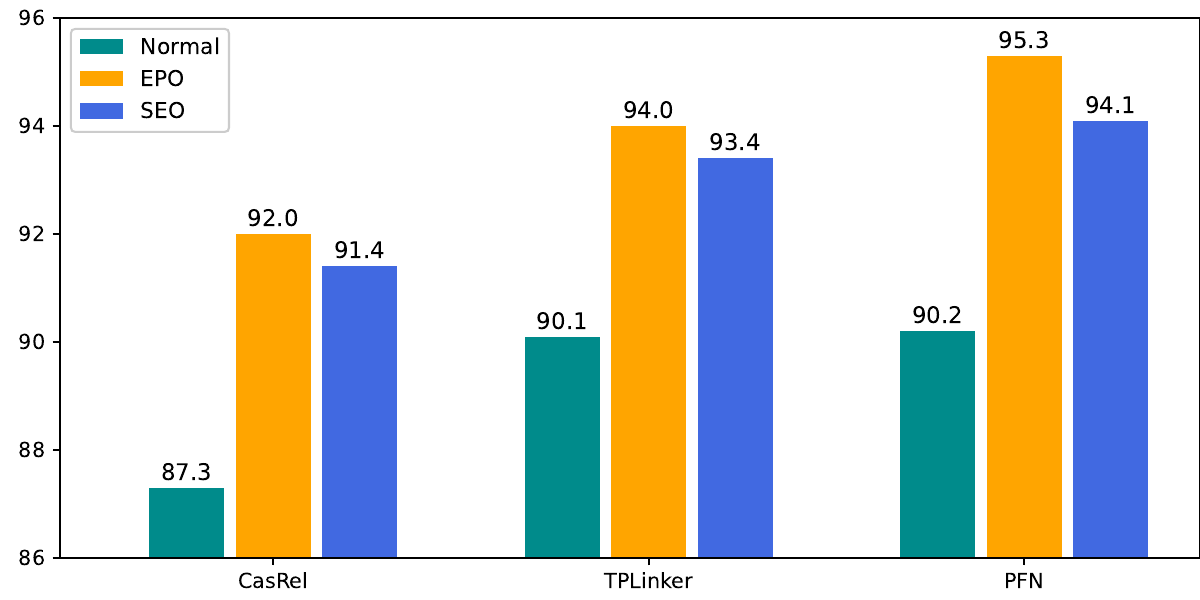}
  }
  \hspace{-0.02\textwidth}
  \subfigure[width=0.49\textwidth][WebNLG]{
  \includegraphics[width=0.49\textwidth]{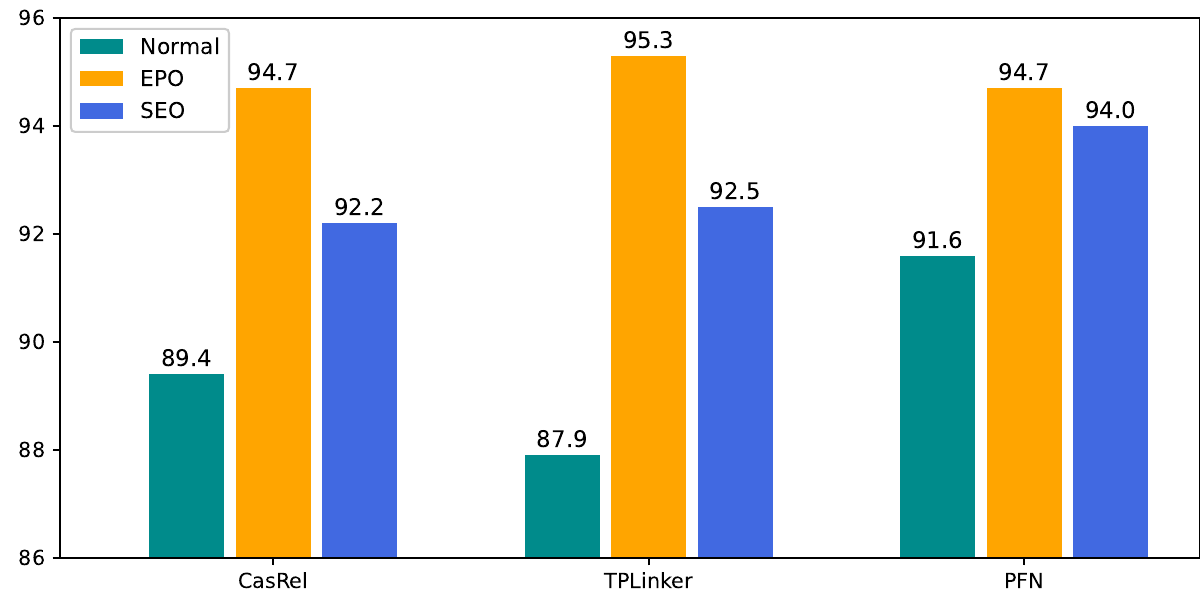}
  }
  \vspace{-0.2cm}
  \caption{\label{fig: normal-epo-spo} F1-score of relation triple extraction on sentences with three different overlapping patterns. }
\end{figure*}

\begin{figure*}[htbp]
 \vspace{-0.2cm}
  \centering
  \subfigure[width=0.49\textwidth][NYT]{
  \includegraphics[width=0.49\textwidth]{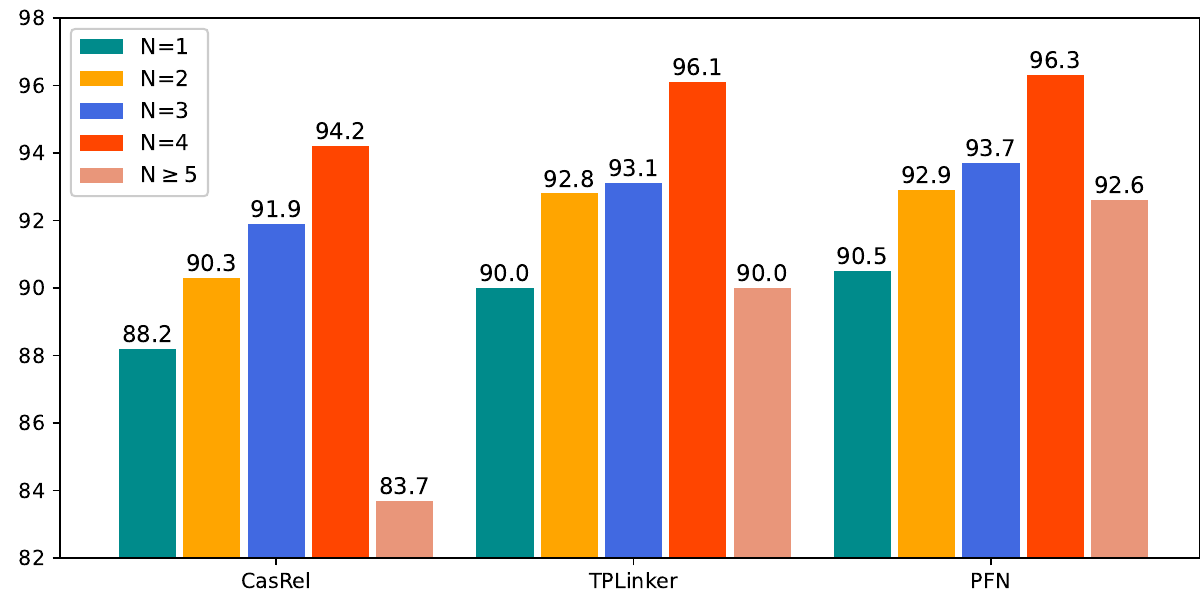}
  }
  \hspace{-0.02\textwidth}
  \subfigure[width=0.49\textwidth][WebNLG]{
  \includegraphics[width=0.49\textwidth]{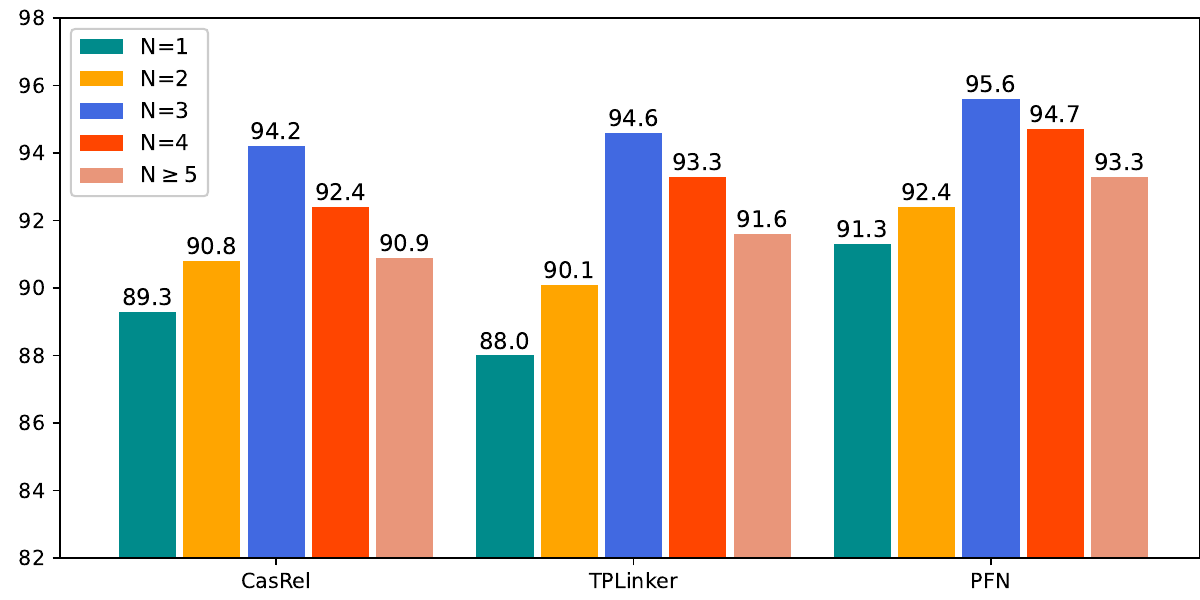}
  }
  \vspace{-0.2cm}
  \caption{\label{fig: nums} F1-score of relational triple extraction on sentences containing \textit{N} triples, with \textit{N} ranges from 1 to $\geq$5.}
  \vspace{-0.2cm}
\end{figure*}

\section{Dataset}
\label{sec A}
We evaluate our model on six datasets. NYT \citep{riedel2010modeling} is sampled from New York Times news articles and annotated by distant supervision. WebNLG is originally created for Natural Language Generation task and is applied by \citep{zeng-etal-2018-extracting} as a relation extraction dataset. ACE05 and ACE04 \citep{walker2006ace} are collected from various sources, including news articles and online forums. ADE \citep{gurulingappa2012development} contains medical descriptions of adverse effects of drug use. SciERC \citep{luan-etal-2018-multi} is collected from 500 AI paper abstracts originally used for scientific knowledge graph construction. Following previous work, we filter out samples containing overlapping entities in ADE, which makes up only 2.8\% of the whole dataset.
Statistics of the datasets can be found in table \ref{tab:my-table1}

\section{Implementation Details}
\label{sec: Appendix c}
We leverage pre-trained language models as our embedding layer. Following previous work, the versions we use are \emph{bert-base-cased}, \emph{albert-xxlarge-v1} and \emph{scibert-scivocab-uncased}. Batch size and learning rate are set to be 4/20 and 1e-5/2e-5 for SciERC/Others respectively. In order to prevent overfitting, dropout \citep{srivastava2014dropout} is used in our word embedding, entity span and triple representation of task units (set to be 0.1). We use Adam \citep{kingma2014adam} to optimize our model parameters and train our model for 100 epochs. Also, to prevent gradient explosion, gradient clipping
is applied during training.

\section{Analysis on Overlapping Pattern and Triple Number}
For more comprehensive evaluation, we assess our model on NYT/WebNLG datasets on different triple overlapping patterns (see section 2 for the detailed description of these patterns) and sentences containing a different number of triples. Since previous work does not compare triple overlapping pattern and triple number in ADE/ACE05/ACE04/ScIERC given that EPO triples are non-existent in these datasets, comparison result is not included for these datasets.

As is shown in figure \ref{fig: normal-epo-spo}, Our model is mostly superior to the other two baselines in all three categories.  Interestingly in normal class, our model performs significantly better in WebNLG, but the score in NYT is basically on par with TpLinker. We argue that this could probably be caused by the fact that NYT, generated by distant supervision, is much more noisier than WebNLG. Besides, sentences of normal triples are likely to be much noisier than sentences of EPO and SEO triples since there is a higher chance for incomplete annotation. Thus it is unsurprising that no significant improvement is achieved in predicting normal triples of NYT. 

Besides, from figure \ref{fig: nums} we observe that our model performs better in sentences with more than five triples on both datasets, where interaction between entity and relation becomes very complex. The strong performance in those sentences confirms the superiority of our model against other baselines.

\section{Details of Robustness Test}
\label{sec B}
Descriptions of the transformation methods used in Table \ref{tab:textflint} are listed as follows:
\begin{enumerate}
\vspace{-2.5mm}
\item ConcatSent - Concatenate sentences to a longer one.
\vspace{-2.5mm}
\item CrossCategory - Entity Swap by swaping entities with ones that can be labeled by different types.
\vspace{-2.5mm}
\item EntTypos - Swap/delete/add random character for entities.
\vspace{-2.5mm}
\item OOV - Entity Swap by out-of-vocabulary entities.
\vspace{-2.5mm}
\item SwapLonger - Substitute short entities for longer ones.
\end{enumerate}

Transformations of RE are not viable for the following reasons:
\begin{enumerate}
\vspace{-2.5mm}
\item The input is restricted to one triple per sentence.
\vspace{-2.5mm}
\item The methods include entity swap, which is already covered in NER.
\vspace{-2.5mm}
\item The methods include relation-specific transformations (Age, Employee, Birth) and ACE05 does not have these type of relations.
\vspace{-2.5mm}
\item The methods include inserting descriptions of entities, which is unfair because it might introduce new entity and relation.
\end{enumerate}


\end{document}